\documentclass[conference]{IEEEtran}
\IEEEoverridecommandlockouts
\usepackage{cite}
\usepackage{amsmath,amssymb,amsfonts}
\usepackage{algorithmic}
\usepackage{graphicx}
\usepackage{textcomp}
\usepackage{xcolor}
\usepackage{algorithm}
\usepackage{booktabs}
\usepackage{multirow}
\usepackage{pifont}
\usepackage{tikz}
\usepackage{pgfplots}
\pgfplotsset{compat=1.18}
\usetikzlibrary{positioning, arrows.meta, shapes.geometric, fit, calc, decorations.pathreplacing}

\def\BibTeX{{\rm B\kern-.05em{\sc i\kern-.025em b}\kern-.08em
    T\kern-.1667em\lower.7ex\hbox{E}\kern-.125emX}}
\begin{document}

\title{Fed-BAC: Federated Bandit-Guided Additive Clustering in Hierarchical Federated Learning\\
\thanks{This work is financed by the European Commission under the Horizon Europe MSCA programme (HORIZON-MSCA-2024-SE-01-01), Grant Agreement No. 101236523 (AeroNet project).}
}

\author{
\IEEEauthorblockN{1\textsuperscript{st} Satwat Bashir }
\IEEEauthorblockA{\textit{Computer Science and Informatics} \\
\textit{London South Bank University}\\
London, UK \\
bashis11@lsbu.ac.uk}
\and
\IEEEauthorblockN{2\textsuperscript{nd} Tasos Dagiuklas }
\IEEEauthorblockA{\textit{Computer Science and Digital Technologies} \\
\textit{London South Bank University}\\
London, UK \\
tdagiuklas@lsbu.ac.uk}

\and
\IEEEauthorblockN{3\textsuperscript{rd} Muddesar Iqbal}
\IEEEauthorblockA{\textit{Computer Science and Digital Technologies} \\
\textit{London South Bank University}\\
London, UK \\
m.iqbal@lsbu.ac.uk}
}

\maketitle

\begin{abstract}
Hierarchical federated learning (HFL) leverages edge servers for partial aggregation in edge computing. Yet existing FL methods lack mechanisms for jointly optimizing cluster assignment and client selection under data heterogeneity. This paper proposes Fed-BAC, which integrates additive cluster personalization with a two-level bandit framework: contextual bandits at the cloud learn server-to-cluster assignments, while Thompson Sampling at each edge server identifies high-contributing clients. The additive decomposition enables the sharing of knowledge between groups through a globally aggregated network, while cluster-specific networks capture distribution variations. Across three classification benchmarks (CIFAR-10, SVHN, Fashion-MNIST) under moderate ($\alpha\!=\!0.5$) and severe ($\alpha\!=\!0.1$) Dirichlet non-IID partitioning, Fed-BAC achieves distributed accuracy gains of up to +35.5pp over HierFAVG and +8.4pp over IFCA, while requiring only 80\% client participation, converging 1.5--4.8$\times$ faster depending on dataset and accuracy target, and improving cross-server fairness. These gains are further validated at $5\times$ deployment scale on CIFAR-10. The advantage of Fed-BAC increases with heterogeneity severity, confirming that additive cluster personalization becomes increasingly valuable as data distributions diverge.
\end{abstract}

\section{Introduction}
\label{sec:introduction}

\IEEEPARstart{F}{ederated} learning (FL) enables distributed devices to collaboratively train a shared model while keeping data local~\cite{mcmahan2017fedavg}. As FL deployments scale to thousands of heterogeneous devices across geographically distributed networks, communication across network tiers becomes a critical bottleneck~\cite{kairouz2021advances, li2020fedprox}.

In real-world deployments, edge servers are an inherent part of the Multi-Access Edge Computing (MEC) infrastructure that mediates between client devices and the cloud. \textit{Hierarchical federated learning} (HFL) leverages this three-tier MEC topology, rather than forcing all client-to-cloud communication through a flat architecture~\cite{liu2020hierfavg, abad2020hierarchical, wu2024topology}. By performing partial aggregation at the edge, HFL reduces client-to-cloud communication rounds~\cite{wang2022demystifying, liu2020hierfavg} and has demonstrated effectiveness across cellular, vehicular, and healthcare verticals. While knowledge distillation~\cite{wu2022communication, zhu2021datafree} and adaptive client selection~\cite{song2023fast} have been explored, the three-tier topology introduces challenges that no single existing method addresses.

Existing HFL methods assume homogeneous data distributions across edge servers, yet real-world deployments exhibit severe statistical heterogeneity~\cite{li2020fedprox, zhao2018federated}. When edge servers aggregate updates from clients with divergent label distributions, the resulting models suffer from client drift~\cite{zhao2018federated, hsieh2020noniid}, with accuracy degradation of up to 20\% compared to IID settings~\cite{hsieh2020noniid}. Clustered FL methods such as IFCA~\cite{ghosh2020ifca} and CFL~\cite{sattler2021cfl} address heterogeneity by grouping similar clients, but existing approaches train fully isolated cluster models with no mechanism for cross-cluster knowledge transfer.

Furthermore, client selection in HFL is predominantly random or static~\cite{mcmahan2017fedavg, cho2022power, lai2021oort}, and existing bandit approaches~\cite{xia2020fedts} operate in flat architectures rather than jointly optimizing client selection and cluster assignment across tiers. Personalization methods such as Ditto~\cite{li2021ditto} and APFL~\cite{deng2020apfl} require $O(N)$ client-level residuals~\cite{tan2022towards}, which the three-tier topology exacerbates by routing residuals through the edge-to-cloud link.

To address these challenges, this paper introduces Fed-BAC, an HFL framework that combines adaptive clustering with learned client selection. Unlike IFCA's isolated clusters, Fed-BAC employs additive cluster personalization building on Fed-CAM~\cite{ma2023fedcam}: predicted logits are decomposed as $\hat{y} = h(x; \Theta_{\mathrm{global}}) + f(x; \Theta_k)$, where a shared global network captures common knowledge while cluster-specific networks capture distribution variations. For adaptive decision-making, Fed-BAC integrates LinUCB~\cite{li2010linucb} at the cloud for cluster assignment and Thompson Sampling (TS)~\cite{russo2020tutorial} at each edge server for client selection. The system architecture is illustrated in Fig.~\ref{fig:architecture}.

The original contributions of this paper are as follows:
\begin{enumerate}
    \item \textbf{Two-Level Bandit Framework for HFL:} A hierarchical decision-making architecture that decouples cluster assignment from client selection. Unlike single-level bandit approaches~\cite{xia2020fedts, cho2020bandit, qu2022context} that target only one decision dimension, Fed-BAC employs LinUCB with four FL-specific contextual features (loss ratio, cluster balance, assignment stability, training progress) at the cloud, and TS at each edge server, enabling different update frequencies suited to each decision's timescale.
    
    \item \textbf{Integrated HFL System with Additive Personalization:} The bandit framework is integrated with logit-space additive cluster personalization~\cite{ma2023fedcam} in a three-tier MEC architecture. Experimental evaluation across three benchmarks under moderate and severe Dirichlet non-IID data demonstrates that Fed-BAC consistently outperforms both HierFAVG and IFCA in distributed accuracy at reduced client participation, with the advantage increasing under more extreme heterogeneity.
\end{enumerate}

\section{Related Work}
\label{sec:related}

\textit{Hierarchical FL.}
Several works have extended FL to hierarchical architectures. Liu et al.~\cite{liu2020hierfavg} introduced HierFAVG, which performs partial aggregation at edge servers before global aggregation at the cloud, demonstrating reduced communication rounds while maintaining competitive accuracy. Abad et al.~\cite{abad2020hierarchical} extended this paradigm to heterogeneous cellular networks with varying server capabilities. However, both approaches assume homogeneous data distributions across edge servers and rely on random client selection. Trindade and da Fonseca~\cite{trindade2024client} addressed client selection in HFL using resource and performance features, but their fixed criteria target only resource heterogeneity and do not incorporate data-distribution-aware clustering or adaptive online learning. More recently, Song et al.~\cite{song2025chpfl} proposed CHPFL, a three-tier framework that uses K-Means++ to cluster clients at the edge level, and Lee et al.~\cite{lee2025phefl} introduced PHE-FL, which personalizes edge models via edge-level knowledge distillation. Both address non-IID data in HFL but rely on static clustering criteria and do not incorporate adaptive online learning for cluster assignment or client selection.

\textit{Clustered FL.}
Clustered FL tackles data heterogeneity by partitioning clients into groups with similar distributions. Ghosh et al.~\cite{ghosh2020ifca} proposed IFCA, which assigns each client to the cluster whose model achieves the lowest loss on local data. Sattler et al.~\cite{sattler2021cfl} and Briggs et al.~\cite{briggs2020federated} introduced gradient-based and hierarchical clustering respectively. All three methods maintain fully isolated per-cluster models. Ma et al.~\cite{ma2023fedcam} addressed this limitation with Fed-CAM, decomposing predictions into a shared global model and cluster-specific residuals, enabling knowledge transfer while preserving specialization. Fed-BAC extends this additive paradigm by replacing static loss-based clustering with bandit-guided dynamic cluster assignment. Licciardi et al.~\cite{licciardi2025fedgwc} recently introduced FedGWC, which clusters clients via a Gaussian-weighted interaction matrix derived from local losses, but operates in flat 2-tier FL without hierarchical aggregation. Ahmad et al.~\cite{ahmad2025cflhkd} proposed CFLHKD, which combines hierarchical CFL with multi-teacher knowledge distillation for inter-cluster knowledge sharing; however, their method does not incorporate adaptive client selection or online learning for cluster assignment.

\textit{Client selection.}
Efficient client selection is critical for reducing communication overhead. Cho et al.~\cite{cho2022power} introduced power-of-choice selection, while Lai et al.~\cite{lai2021oort} proposed Oort, which combines statistical and system utility for guided participant selection. Bandit-based approaches~\cite{xia2020fedts, cho2020bandit} formulate client selection as an online learning problem and have demonstrated promising results in flat architectures. Qu et al.~\cite{qu2022context} extended this direction to hierarchical settings with context-aware multi-armed bandits. Despite these advances, existing bandit methods operate at a single decision level and do not jointly optimize cluster assignment and client selection across different network tiers.

\textit{Personalized FL.}
Client-level personalization methods such as Ditto~\cite{li2021ditto}, APFL~\cite{deng2020apfl}, and FedRep~\cite{collins2021fedrep} maintain per-client personalized components, requiring $O(N)$ additional parameters that become impractical at scale. Table~\ref{tab:related_work} summarizes the main characteristics of existing approaches and shows that no existing method jointly addresses hierarchical aggregation, adaptive clustering with cross-cluster knowledge sharing, and learned client selection. Fed-BAC bridges this gap by combining hierarchical bandits at multiple decision levels with additive clustered personalization in HFL.

\begin{table*}[t]
\centering
\caption{Related work comparison. ``Adaptive'' = online learning during training.}
\label{tab:related_work}
\begin{tabular}{lccccccc}
\toprule
\textbf{Reference} & \textbf{Year} & \textbf{Architecture} & \textbf{Clustering} & \textbf{Client Selection} & \textbf{Adaptive} & \textbf{Knowledge Sharing} & \textbf{Personalization} \\
\midrule
Liu et al.~\cite{liu2020hierfavg} & 2020 & 3-tier & \ding{55} & Random & \ding{55} & Full & \ding{55} \\
Abad et al.~\cite{abad2020hierarchical} & 2020 & 3-tier & \ding{55} & Random & \ding{55} & Full & \ding{55} \\
Trindade \& Fonseca~\cite{trindade2024client} & 2024 & 3-tier & \ding{55} & Resource-aware & \ding{51} & Full & \ding{55} \\
Song et al.~\cite{song2025chpfl} & 2025 & 3-tier & K-Means++ & Random & \ding{55} & Full & Cluster \\
Lee et al.~\cite{lee2025phefl} & 2025 & 3-tier & \ding{55} & Random & \ding{55} & Distillation & Edge \\
\midrule
Ghosh et al.~\cite{ghosh2020ifca} & 2020 & 2-tier & Loss-based & Random & \ding{55} & None & Cluster \\
Sattler et al.~\cite{sattler2021cfl} & 2021 & 2-tier & Gradient-based & Random & \ding{55} & None & Cluster \\
Briggs et al.~\cite{briggs2020federated} & 2020 & 2-tier & Hierarchical & Random & \ding{55} & None & Cluster \\
Ma et al.~\cite{ma2023fedcam} & 2023 & 2-tier & Loss-based & Random & \ding{55} & Full (additive) & Cluster \\
Licciardi et al.~\cite{licciardi2025fedgwc} & 2025 & 2-tier & Gaussian-weighted & Random & \ding{55} & None & Cluster \\
\midrule
Cho et al.~\cite{cho2022power} & 2022 & 2-tier & \ding{55} & Power-of-choice & \ding{55} & Full & \ding{55} \\
Lai et al.~\cite{lai2021oort} & 2021 & 2-tier & \ding{55} & Utility-based & \ding{51} & Full & \ding{55} \\
Xia et al.~\cite{xia2020fedts} & 2020 & 2-tier & \ding{55} & MAB & \ding{51} & Full & \ding{55} \\
Qu et al.~\cite{qu2022context} & 2022 & 3-tier & \ding{55} & CC-MAB & \ding{51} & Full & \ding{55} \\
Lu et al.~\cite{lu2026fedcsad} & 2026 & 3-tier & \ding{55} & Contextual bandit & \ding{51} & Distillation & Edge \\
\midrule
Li et al.~\cite{li2021ditto} & 2021 & 2-tier & \ding{55} & Random & \ding{55} & Regularization & Client \\
Collins et al.~\cite{collins2021fedrep} & 2021 & 2-tier & \ding{55} & Random & \ding{55} & Shared layers & Client \\
\midrule
\textbf{Fed-BAC (Proposed)} & \textbf{2026} & \textbf{3-tier} & \textbf{LinUCB} & \textbf{Thompson Sampling} & \textbf{\ding{51}} & \textbf{Full (additive)} & \textbf{Cluster} \\
\bottomrule
\end{tabular}
\end{table*}

\section{System Model and Problem Formulation}
\label{sec:system_model}

\subsection{System Architecture}

This work considers a three-tier MEC-based HFL system consisting of a cloud server, $M$ edge servers, and $N$ clients distributed across the edge servers. Each edge server $m \in \{1, \ldots, M\}$ manages a set of clients $S_m$ with $|S_m| = N_m$, giving a total of $N = \sum_{m=1}^{M} N_m$ clients. Each client $i$ holds a private local dataset $D_i = \{(x_{i,j}, y_{i,j})\}_{j=1}^{n_i}$ that remains on-device throughout training.

The cloud server maintains $K_{\max}$ cluster models. Since servers are the units being clustered, $K_{\max} \leq M$ is a natural upper bound (in all experiments, $K_{\max} = M$). The number of \textit{active} clusters $K^{(t)} \leq K_{\max}$ is determined by the LinUCB bandit during training. The cluster assignment function $\pi^{(t)}\colon \{1, \ldots, M\} \to \{1, \ldots, K_{\max}\}$ maps each edge server to a cluster at round~$t$, inducing the partition $\mathcal{C}_k^{(t)} = \{m : \pi^{(t)}(m) = k\}$, and is learned adaptively via LinUCB so that assignments evolve during training.

Training proceeds in synchronous rounds $t = 1, \ldots, T$. Each round consists of six phases:
\begin{enumerate}
    \item \textbf{Model Distribution:} The cloud transmits $\Theta_{\mathrm{global}}^{(t)}$ and $\Theta_{\pi^{(t)}(m)}^{(t)}$ to each edge server~$m$.
    
    \item \textbf{Client Selection:} Each edge server~$m$ selects a subset $\mathcal{S}_m^{(t)} \subseteq S_m$ with $|\mathcal{S}_m^{(t)}| = B$ via TS, where $B = \lfloor p \cdot N_m \rfloor$ is the per-round budget at participation rate $p \in (0, 1]$.
    
    \item \textbf{Local Training:} Each selected client $i \in \mathcal{S}_m^{(t)}$ performs $E$ epochs of SGD on $D_i$, jointly training both networks.
    
    \item \textbf{Edge Aggregation:} Each edge server~$m$ computes:
    \begin{equation}
    \label{eq:edge_agg}
    (\bar{\Theta}_{\mathrm{global},m}^{(t)},\, \bar{\Theta}_{k,m}^{(t)}) = \sum_{i \in \mathcal{S}_m^{(t)}} \frac{n_i}{n_{\mathcal{S}_m}} \, (\Theta_{\mathrm{global},i}^{(t)},\, \Theta_{k,i}^{(t)}),
    \end{equation}
    where $n_{\mathcal{S}_m} = \sum_{i \in \mathcal{S}_m^{(t)}} n_i$ is the total data size of the selected clients.
    
    \item \textbf{Cloud Aggregation:} Two-phase aggregation updates global and cluster networks (Section~\ref{sec:additive}).
    
    \item \textbf{Cluster Reassignment:} Every $\tau_{\mathrm{re}}$ rounds, LinUCB re-evaluates $\pi^{(t)}(m)$ for each server.
\end{enumerate}

\subsection{Problem Formulation}

Let $\ell(\theta; x, y)$ denote the loss for sample $(x, y)$ under parameters $\theta$. The local objective for client~$i$ is $F_i(\theta) = \frac{1}{n_i} \sum_{j=1}^{n_i} \ell(\theta;\, x_{i,j},\, y_{i,j})$, and the edge server objective is $F_m(\theta) = \sum_{i \in S_m} \frac{n_i}{n_m}\, F_i(\theta)$, where $n_m = \sum_{i \in S_m} n_i$ is the total data size at server~$m$ and $n = \sum_{m=1}^{M} n_m$ is the total data size across all clients.

Fed-BAC adopts an additive decomposition in the logit space following Fed-CAM~\cite{ma2023fedcam}. The predicted logits for a client in cluster~$k$ are:
\begin{equation}
\label{eq:additive}
\hat{y} = h(x;\, \Theta_{\mathrm{global}}) + f(x;\, \Theta_k),
\end{equation}
where $h(\cdot;\, \Theta_{\mathrm{global}})$ is a global network capturing shared knowledge and $f(\cdot;\, \Theta_k)$ is a cluster-specific network capturing distribution deviations. Under this decomposition, the model parameters for cluster~$k$ are the pair $(\Theta_{\mathrm{global}}, \Theta_k)$, and $F_m(\Theta_{\mathrm{global}}, \Theta_k)$ denotes evaluation under Eq.~\eqref{eq:additive}.

The joint optimization problem is:
\begin{equation}
\label{eq:joint_opt}
\min_{\Theta_{\mathrm{global}},\, \{\Theta_k\}_{k=1}^{K_{\max}},\, \pi} \;\;
\sum_{k=1}^{K_{\max}} \sum_{m \in \mathcal{C}_k} \frac{n_m}{n}\, F_m(\Theta_{\mathrm{global}}, \Theta_k),
\end{equation}
subject to $\pi$ being a valid partition of $\{1, \ldots, M\}$ into at most $K_{\max}$ clusters, and $|\mathcal{S}_m^{(t)}| = B$ for all servers and rounds. Since $\pi$ and $\{\Theta_k\}$ are interdependent, this objective is solved iteratively: model parameters are updated each round via SGD, while $\pi$ and client selection are updated via online learning. This presents three coupled subproblems: (1)~model optimization given fixed assignments and selected clients, (2)~cluster assignment given current models, and (3)~client selection given models and assignments. IFCA~\cite{ghosh2020ifca} addresses~(2) via alternating minimization with static loss-based assignment and ignores~(3) entirely. Fed-BAC employs online learning for both: LinUCB for cluster assignment and TS for client selection.

\subsection{Communication Cost Model}

Let $d_g$ and $d_k$ denote the global and cluster network sizes in Fed-BAC's additive decomposition. In Fed-BAC, each client downloads and uploads both networks, so client-to-edge communication per round is:
\begin{equation}
\label{eq:comm_cost}
C_{\mathrm{client}}^{(t)} = 2(d_g + d_k) \sum_{m=1}^{M} |\mathcal{S}_m^{(t)}| = 2p N (d_g + d_k).
\end{equation}
The second equality holds when $p \cdot N_m$ is integer for all servers (as in the experimental setup with $p\!=\!0.8$, $N_m\!=\!10$). Since each participating client transmits both networks ($d_g + d_k \approx 2d_{\text{model}}$), the per-round byte cost at the client-to-edge tier is $0.8 \times 2 = 1.6\times$ that of full-participation single-model baselines. Fed-BAC therefore does not reduce per-round communication volume; instead, fewer active devices per round lower scheduling overhead, energy consumption, and wireless contention in MEC deployments~\cite{trindade2024client}. However, faster convergence (Section~\ref{sec:convergence}) offsets the higher per-round cost: on CIFAR-10 at $\alpha\!=\!0.5$, Fed-BAC reaches 50\% accuracy by round~15 versus round~72 for HierFAVG, yielding $15 \times 1.6 = 24$ normalized round-equivalents versus~72, a $3.0\times$ total-byte reduction. At $\alpha\!=\!0.1$ the ratio improves to $5.9\times$. McMahan et~al.~\cite{mcmahan2017fedavg} demonstrated that partial participation can reduce total rounds to converge; Fed-BAC extends this to a three-tier setting where convergence speedups more than compensate for the doubled model payload.

\subsection{Data Heterogeneity Model}

Non-IID data is modeled using two-level Dirichlet partitioning~\cite{hsu2019measuring}. Let $C$ denote the number of classes. Server-level heterogeneity: $\mathbf{q}_m^{\mathrm{server}} \sim \mathrm{Dir}(\alpha_{\mathrm{server}} \cdot \mathbf{1}_C)$. Client-level heterogeneity within each server~$m$: $\mathbf{q}_i^{\mathrm{client}} \sim \mathrm{Dir}(\alpha_{\mathrm{client}} \cdot \mathbf{1}_C)$, $i \in S_m$. For each class, training samples are allocated across servers (or clients within a server) by drawing counts from a multinomial distribution parameterized by the corresponding Dirichlet-sampled proportions. Experiments evaluate two settings: moderate heterogeneity ($\alpha_{\mathrm{server}} = \alpha_{\mathrm{client}} = 0.5$) and severe heterogeneity ($\alpha_{\mathrm{server}} = 0.1$, $\alpha_{\mathrm{client}} = 0.5$), where the latter produces highly skewed server-level distributions while preserving within-server variation.

\section{Proposed Fed-BAC Framework}
\label{sec:framework}

\begin{figure*}[t]
\centering
\begin{tikzpicture}[
    node distance=1.2cm and 1.0cm,
    cloud/.style={draw, ellipse, minimum width=2.8cm, minimum height=1.2cm, fill=blue!8, thick},
    server/.style={draw, rectangle, rounded corners=3pt, minimum width=1.6cm, minimum height=0.8cm, fill=green!8, thick},
    client/.style={draw, circle, minimum size=0.55cm, fill=orange!12, thick, font=\scriptsize},
    arr/.style={-{Stealth[length=2.5mm]}, thick},
    darr/.style={{Stealth[length=2.5mm]}-{Stealth[length=2.5mm]}, thick},
    lbl/.style={font=\scriptsize, midway},
]

\node[cloud] (cloud) {\begin{tabular}{c}\textbf{Cloud Server}\\[-2pt]{\scriptsize $\Theta_{\mathrm{global}},\;\{\Theta_k\}_{k=1}^{K_{\max}}$}\end{tabular}};

\node[above=0.15cm of cloud, font=\scriptsize\itshape, text=blue!70!black] {Per-server LinUCB bandits};

\node[server, below left=1.4cm and 3.2cm of cloud] (s1) {\begin{tabular}{c}\scriptsize Server 1\\[-2pt]{\tiny Cluster $k_1$}\end{tabular}};
\node[server, below left=1.4cm and 1.0cm of cloud] (s2) {\begin{tabular}{c}\scriptsize Server 2\\[-2pt]{\tiny Cluster $k_2$}\end{tabular}};
\node[below=1.4cm of cloud, font=\large] (dots) {$\cdots$};
\node[server, below right=1.4cm and 1.0cm of cloud] (s9) {\begin{tabular}{c}\scriptsize Server $M\!-\!1$\\[-2pt]{\tiny Cluster $k_{M-1}$}\end{tabular}};
\node[server, below right=1.4cm and 3.2cm of cloud] (s10) {\begin{tabular}{c}\scriptsize Server $M$\\[-2pt]{\tiny Cluster $k_M$}\end{tabular}};

\node[below=0.05cm of s1, font=\tiny\itshape, text=red!60!black] {Thompson Sampling};
\node[below=0.05cm of s10, font=\tiny\itshape, text=red!60!black] {Thompson Sampling};

\draw[darr, blue!60] (cloud) -- (s1) node[lbl, left, text=black] {\tiny $(\Theta_{\mathrm{global}}, \Theta_{k_1})$};
\draw[darr, blue!60] (cloud) -- (s2);
\draw[darr, blue!60] (cloud) -- (s9);
\draw[darr, blue!60] (cloud) -- (s10) node[lbl, right, text=black] {\tiny $(\Theta_{\mathrm{global}}, \Theta_{k_M})$};

\foreach \i in {1,2,3} {
    \node[client, below=0.9cm of s1, xshift={(\i-2)*0.7cm}] (c1\i) {\i};
}
\node[below=0.9cm of s1, xshift=1.4cm, font=\tiny] {$\cdots$};
\node[client, below=0.9cm of s1, xshift=1.9cm] (c1n) {\tiny $N_1$};

\foreach \i in {1,2,3} {
    \node[client, below=0.9cm of s10, xshift={(\i-2)*0.7cm}] (c10\i) {\i};
}
\node[below=0.9cm of s10, xshift=1.4cm, font=\tiny] {$\cdots$};
\node[client, below=0.9cm of s10, xshift=1.9cm] (c10n) {\tiny $N_M$};

\draw[arr, orange!70!black] (s1) -- (c11);
\draw[arr, orange!70!black] (s1) -- (c12);
\draw[arr, orange!70!black, dashed, thin] (s1) -- (c13);
\draw[arr, orange!70!black] (s1) -- (c1n);

\draw[arr, orange!70!black] (s10) -- (c101);
\draw[arr, orange!70!black, dashed, thin] (s10) -- (c102);
\draw[arr, orange!70!black] (s10) -- (c103);
\draw[arr, orange!70!black] (s10) -- (c10n);

\node[draw, rounded corners, fill=gray!5, inner sep=4pt, font=\scriptsize] at ($(cloud)+(6.5,-1.0)$) {
\begin{tabular}{ll}
\tikz\draw[darr,blue!60] (0,0) -- (0.6,0); & Cloud $\leftrightarrow$ Server \\
\tikz\draw[arr,orange!70!black] (0,0) -- (0.6,0); & Selected client \\
\tikz\draw[arr,orange!70!black,dashed,thin] (0,0) -- (0.6,0); & Unselected client \\
\end{tabular}
};

\node[left=0.3cm of cloud, font=\scriptsize, text=gray] {\textit{Cloud tier}};
\node[left=0.3cm of s1, font=\scriptsize, text=gray] {\textit{Edge tier}};
\node[left=0.3cm of c11, font=\scriptsize, text=gray] {\textit{Client tier}};

\end{tikzpicture}
\caption{Fed-BAC system architecture. LinUCB at the cloud assigns servers to clusters; TS at each edge server selects $B$ of $N_m$ clients per round.}
\label{fig:architecture}
\end{figure*}
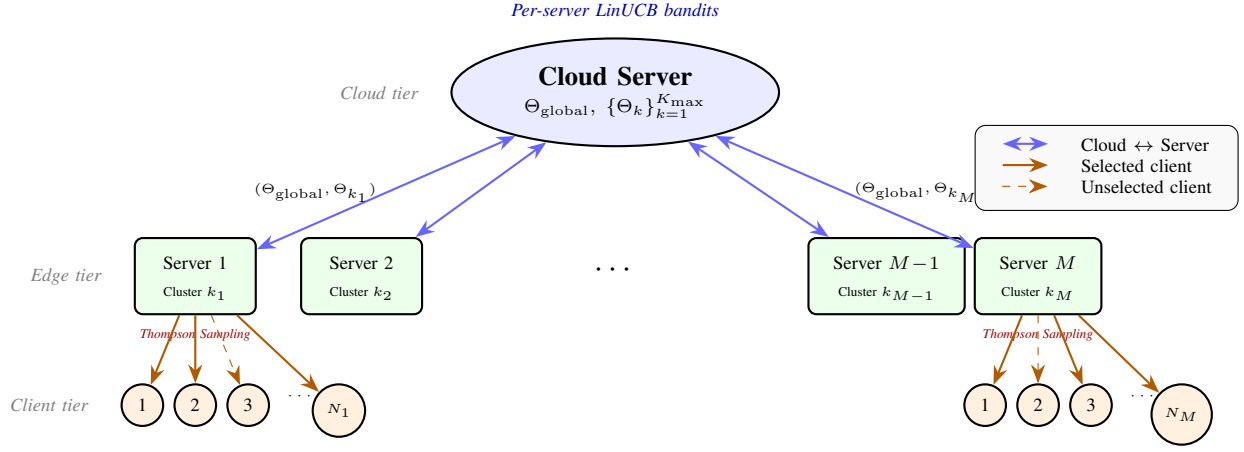

The Fed-BAC framework, illustrated in Fig.~\ref{fig:architecture}, comprises additive cluster personalization and a two-level bandit integration.

\subsection{Additive Cluster Personalization}
\label{sec:additive}

Building on the additive decomposition introduced in Eq.~\eqref{eq:additive}, Fed-BAC differs from Fed-CAM~\cite{ma2023fedcam} in two respects: static loss-based clustering is replaced with learned bandit-guided assignment, and learned client selection is incorporated at the edge level.

\subsubsection{Two-Phase Aggregation}

\textit{Phase~1 (Global Aggregation):} The cloud aggregates from \textit{all} $M$ edge servers:
\begin{equation}
\label{eq:global_agg}
\Theta_{\mathrm{global}}^{(t+1)} = \sum_{m=1}^{M} \frac{n_m}{n}\, \bar{\Theta}_{\mathrm{global},m}^{(t)}.
\end{equation}
Weights use total server data sizes $n_m$ (not selected-subset sizes $n_{\mathcal{S}_m}$) to maintain stable aggregation weights independent of the stochastic client selection, following standard FedAvg practice~\cite{mcmahan2017fedavg}.

\textit{Phase~2 (Cluster Aggregation):} Within each cluster:
\begin{equation}
\label{eq:residual}
\Theta_k^{(t+1)} = \sum_{m \in \mathcal{C}_k^{(t)}} \frac{n_m}{n_k}\, \bar{\Theta}_{k,m}^{(t)},
\end{equation}
where $n_k = \sum_{m \in \mathcal{C}_k^{(t)}} n_m$ is the total data size of cluster~$k$, with additional L2 regularization $\lambda = 0.001$ applied specifically to cluster parameters $\Theta_k$ during local training (separate from the general weight decay in Section~\ref{sec:experiments}). Unlike IFCA which aggregates only within clusters, all $M$ servers contribute to $\Theta_{\mathrm{global}}$, enabling cross-cluster knowledge sharing.

\subsection{Two-Level Bandit Integration}

\subsubsection{LinUCB for Cluster Assignment}

At the cloud level, independent per-server LinUCB~\cite{li2010linucb} bandits are employed to prevent herd behavior. Let $L_m^{(k)} = F_m(\Theta_{\mathrm{global}}, \Theta_k)$ denote the loss of cluster~$k$'s model on server~$m$'s data, and $\bar{k}_m \triangleq \arg\min_{k \neq k_m} L_m^{(k)}$ the best alternative cluster. For each server~$m$, a contextual feature vector $\mathbf{x}_m \in \mathbb{R}^4$ is extracted:
\begin{equation}
\label{eq:features}
\mathbf{x}_m = \begin{bmatrix}
\ln\!\dfrac{L_m^{(k_m)} + \epsilon}{L_m^{(\bar{k}_m)} + \epsilon} \\[8pt]
\displaystyle\frac{|\mathcal{C}_{k_m}| - |\mathcal{C}_{\bar{k}_m}|}{|\mathcal{C}_{k_m}| + |\mathcal{C}_{\bar{k}_m}|} \\[6pt]
\min\!\bigl(\tau_m / 2\tau_{\mathrm{re}},\; 1\bigr) \\[4pt]
t \,/\, T
\end{bmatrix},
\end{equation}
where $\tau_m$ denotes the number of consecutive rounds server~$m$ has remained in its current cluster. The four features capture relative cluster fit, cluster balance, assignment stability, and training phase, respectively; the first feature is negative when the current cluster fits better, and the third saturates after $2\tau_{\mathrm{re}}$ rounds.

Each server's cluster assignment is:
\begin{equation}
\label{eq:linucb}
k_m^* = \arg\max_{k \in \{1,\ldots,K_{\max}\}} \left[ \hat{\boldsymbol{\theta}}_k^{(m)\top} \mathbf{x}_m \;+\; \alpha_{\mathrm{UCB}} \sqrt{\mathbf{x}_m^\top (\mathbf{A}_k^{(m)})^{-1} \mathbf{x}_m} \right],
\end{equation}
where $\hat{\boldsymbol{\theta}}_k^{(m)} = (\mathbf{A}_k^{(m)})^{-1} \mathbf{b}_k^{(m)}$ and $\alpha_{\mathrm{UCB}} = 0.3$ controls exploration, set empirically to balance early cluster exploration against exploitation of learned assignments. The reward uses a normalized loss ratio:
\begin{equation}
\label{eq:reward}
r_m = \frac{L_m^{(\bar{k}_m)} - L_m^{(k_m)}}{L_m^{(\bar{k}_m)} + L_m^{(k_m)} + \epsilon},
\end{equation}
bounded in $[-1, 1]$, evaluated every $\tau_{\mathrm{re}} = 20$ rounds, with $\epsilon = 10^{-8}$ for numerical stability. A positive reward indicates that the current cluster provides lower loss than the best alternative, serving as a heuristic proxy for the per-server contribution to the global objective in Eq.~\eqref{eq:joint_opt}. After observing $r_m$, the bandit parameters for the \textit{current} assignment $k_m$ are updated:
\begin{equation}
\label{eq:linucb_update}
\mathbf{A}_{k_m}^{(m)} \leftarrow \mathbf{A}_{k_m}^{(m)} + \mathbf{x}_m \mathbf{x}_m^\top, \quad
\mathbf{b}_{k_m}^{(m)} \leftarrow \mathbf{b}_{k_m}^{(m)} + r_m \,\mathbf{x}_m.
\end{equation}
The new assignment $k_m^*$ is then selected via Eq.~\eqref{eq:linucb} using the updated parameters.

\subsubsection{Thompson Sampling for Client Selection}

At the edge level, each server uses TS~\cite{russo2020tutorial}. For each client~$i$, a $\mathrm{Beta}(\alpha_i, \beta_i)$ prior is maintained, initialized as $\mathrm{Beta}(1,1)$ (non-informative uniform prior). Each round: (1)~sample $\tilde{p}_i \sim \mathrm{Beta}(\alpha_i, \beta_i)$, (2)~select top $B$ clients, (3)~update posteriors using soft updates:
\begin{equation}
\label{eq:ts_magnitude}
\Delta = \min(10 \cdot |r^{\mathrm{TS}}|,\; 2.0),
\end{equation}
\begin{equation}
\label{eq:ts_update}
\alpha_i \leftarrow \alpha_i + \Delta \cdot \mathbb{1}[r^{\mathrm{TS}} > 0], \quad
\beta_i \leftarrow \beta_i + \Delta \cdot \mathbb{1}[r^{\mathrm{TS}} \leq 0],
\end{equation}
where $r^{\mathrm{TS}} = a_m^{(t)} - a_m^{(t-1)}$ is the single-round accuracy change (distinct from the LinUCB reward $r_m$ in Eq.~\eqref{eq:reward}), with $a_m^{(t)}$ denoting the classification accuracy of server~$m$'s cluster model on its local test data at round~$t$. All selected clients receive the same collective reward; individual contributions are distinguished over time through repeated stochastic sampling. Posteriors for unselected clients remain unchanged. Random selection is used during the first $\tau_{\mathrm{TS}} = 10$ warmup rounds. The discriminative power of TS at high participation rates is analyzed in Section~\ref{sec:client_selection}.

\subsection{Algorithm Summary and Analysis}

Algorithms~\ref{alg:cloud} and~\ref{alg:edge} present the cloud orchestration and edge server procedures, respectively. Algorithm~\ref{alg:cloud} governs the cloud server: at each round~$t$, it distributes the current global and cluster models to all edge servers, collects their locally aggregated updates via Algorithm~\ref{alg:edge}, performs two-phase aggregation (global in line~8, then per-cluster in lines~9--11), and every $\tau_{\mathrm{re}}$ rounds triggers LinUCB reassignment (lines~12--18) by extracting contextual features, computing rewards, and updating cluster assignments. Algorithm~\ref{alg:edge} operates at each edge server: it selects $B$ clients via TS (or randomly during the first $\tau_{\mathrm{TS}}$ warmup rounds), dispatches local SGD training in parallel, computes weighted edge aggregation, evaluates the resulting model on local test data, and updates the TS posteriors based on the single-round accuracy change.

\begin{algorithm}[t]
\caption{Fed-BAC: Cloud Orchestration}
\label{alg:cloud}
\begin{algorithmic}[1]
\REQUIRE $\Theta_{\mathrm{global}}^{(0)}$, $K_{\max}$, $T$, $B$, $\tau_{\mathrm{re}}$, $\tau_{\mathrm{TS}}$
\STATE Init $\Theta_{\mathrm{global}} \!\leftarrow\! \Theta_{\mathrm{global}}^{(0)}$;\; $\Theta_k \!\leftarrow\! \Theta_k^{(0)}$,\; $\mathbf{A}_k^{(m)} \!\leftarrow\! \mathbf{I}_4$,\; $\mathbf{b}_k^{(m)} \!\leftarrow\! \mathbf{0}$ \;$\forall m,k$
\STATE Init $\alpha_i \!\leftarrow\! 1$, $\beta_i \!\leftarrow\! 1$ $\forall i$;\; assignments $\{k_m\}_{m=1}^{M}$
\FOR{$t = 1$ to $T$}
    \STATE Send $(\Theta_{\mathrm{global}},\, \Theta_{k_m})$ to each server $m$
    \FOR{each server $m$ \textbf{in parallel}}
        \STATE $(\bar{\Theta}_{\mathrm{global},m}, \bar{\Theta}_{k,m}) \leftarrow \text{\textsc{EdgeRound}}(m, t)$ \hfill \textit{// Alg.~\ref{alg:edge}}
    \ENDFOR
    \STATE $\Theta_{\mathrm{global}} \leftarrow \sum_{m=1}^{M} \frac{n_m}{n}\, \bar{\Theta}_{\mathrm{global},m}$ \hfill \textit{// Global agg.}
    \FOR{$k = 1$ to $K_{\max}$ \textbf{where} $\mathcal{C}_k \neq \emptyset$}
        \STATE $\Theta_k \leftarrow \sum_{m \in \mathcal{C}_k} \frac{n_m}{n_k}\, \bar{\Theta}_{k,m}$ \hfill \textit{// Cluster agg.}
    \ENDFOR
    \IF{$t \bmod \tau_{\mathrm{re}} = 0$}
        \FOR{$m = 1, \ldots, M$}
            \STATE $\mathbf{x}_m \!\leftarrow\! \text{Eq.~\eqref{eq:features}}$;\; $r_m \!\leftarrow\! \text{Eq.~\eqref{eq:reward}}$
            \STATE Update $\mathbf{A}_{k_m}^{(m)}\!, \mathbf{b}_{k_m}^{(m)}$ via Eq.~\eqref{eq:linucb_update}
            \STATE $k_m \leftarrow \arg\max_k \mathrm{UCB}_k^{(m)}(\mathbf{x}_m)$ via Eq.~\eqref{eq:linucb}
        \ENDFOR
    \ENDIF
\ENDFOR
\RETURN $\Theta_{\mathrm{global}}, \{\Theta_k\}_{k=1}^{K_{\max}}$
\end{algorithmic}
\end{algorithm}

\begin{algorithm}[t]
\caption{\textsc{EdgeRound}: Edge Server $m$ at Round $t$}
\label{alg:edge}
\begin{algorithmic}[1]
\IF{$t \leq \tau_{\mathrm{TS}}$}
    \STATE $\mathcal{S}_m^{(t)} \leftarrow$ random $B$ clients from $S_m$ \hfill \textit{// Warmup}
\ELSE
    \STATE Sample $\tilde{p}_i \!\sim\! \mathrm{Beta}(\alpha_i, \beta_i)$ $\forall i \!\in\! S_m$;\; $\mathcal{S}_m^{(t)} \!\leftarrow$ top-$B$
\ENDIF
\FOR{each $i \in \mathcal{S}_m^{(t)}$ \textbf{in parallel}}
    \STATE $(\Theta_{\mathrm{global},i}, \Theta_{k,i}) \leftarrow \mathrm{LocalSGD}((\Theta_{\mathrm{global}}, \Theta_{k_m}), D_i, E)$
\ENDFOR
\STATE $(\bar{\Theta}_{\mathrm{global},m}, \bar{\Theta}_{k,m}) \leftarrow \sum_{i \in \mathcal{S}_m^{(t)}} \frac{n_i}{n_{\mathcal{S}_m}} (\Theta_{\mathrm{global},i}, \Theta_{k,i})$
\STATE $a_m^{(t)} \!\leftarrow\! \mathrm{Acc}(\bar{\Theta}_{\mathrm{global},m}, \bar{\Theta}_{k,m};\, D_m^{\mathrm{test}})$
\STATE $r^{\mathrm{TS}} \!\leftarrow\! a_m^{(t)} \!-\! a_m^{(t-1)}$;\; update $\alpha_i, \beta_i$ $\forall i \!\in\! \mathcal{S}_m^{(t)}$ via Eqs.~\eqref{eq:ts_magnitude}--\eqref{eq:ts_update}
\RETURN $(\bar{\Theta}_{\mathrm{global},m}, \bar{\Theta}_{k,m})$
\end{algorithmic}
\end{algorithm}

\textit{Complexity.} LinUCB requires $O(d^3 MK_{\max})$ operations every $\tau_{\mathrm{re}}$ rounds (e.g. $d\!=\!4$, yielding $O(64MK_{\max})$). The dominant cost is $M \cdot K_{\max}$ forward passes for cluster evaluation, amortized over $\tau_{\mathrm{re}}\!=\!20$ rounds. TS requires $O(N_m)$ beta samples per server per round.

\textit{Convergence.} The individual components have well-studied guarantees: FedAvg-style aggregation~\cite{mcmahan2017fedavg,li2020fedprox}, LinUCB with $O(\sqrt{T \log T})$ regret~\cite{li2010linucb}, and TS with near-optimal regret~\cite{agrawal2012thompson}. However, these assume stationary rewards; in Fed-BAC, rewards are non-stationary as models improve, so standard bounds do not directly apply. The experimental results in Section~\ref{sec:experiments} provide empirical convergence evidence across all benchmarks.

\section{Experimental Evaluation}
\label{sec:experiments}

\subsection{Experimental Setup}

\subsubsection{Datasets and Architecture}
The experiments adopt a three-tier MEC hierarchy with $M\!=\!10$ edge servers and $N\!=\!100$ clients ($N_m\!=\!10$) as the primary configuration. To validate that the gains hold at larger deployment scale, an additional CIFAR-10 evaluation is conducted at $5\times$ client population ($M\!=\!20$, $N_m\!=\!25$, total $N\!=\!500$) under the same training protocol, and is reported in Section~\ref{sec:results}. To evaluate robustness to heterogeneity, two Dirichlet configurations are tested at the primary scale: moderate ($\alpha_{\mathrm{server}} = \alpha_{\mathrm{client}} = 0.5$) and severe ($\alpha_{\mathrm{server}} = 0.1$, $\alpha_{\mathrm{client}} = 0.5$); the scaling study uses the moderate setting to match the headline configuration of Table~\ref{tab:main_results}. Experiments use three datasets:
\begin{enumerate}
    \item \textbf{CIFAR-10}~\cite{krizhevsky2009cifar}: 50K/10K train/test, 10 classes, $32\!\times\!32$ RGB, $T\!=\!200$ rounds.
    \item \textbf{SVHN}~\cite{netzer2011svhn}: 73K/26K train/test, 10 classes, $32\!\times\!32$ RGB, $T\!=\!150$ rounds.
    \item \textbf{Fashion-MNIST}~\cite{xiao2017fmnist}: 60K/10K train/test, 10 classes, $28\!\times\!28$ grayscale, $T\!=\!100$ rounds.
\end{enumerate}

\subsubsection{Model and Training}
All methods use LeNet-5 (approximately 62K parameters for RGB, approximately 44K for grayscale). For Fed-BAC, both global and cluster networks use this architecture, yielding approximately 124K total parameters for RGB datasets (${\approx}\,2\times$ the baseline capacity). This additional capacity is inherent to the additive decomposition, where $\Theta_{\mathrm{global}}$ and $\Theta_k$ serve complementary roles; Section~\ref{sec:client_selection} isolates the bandit contribution at identical capacity, and Section~\ref{sec:discussion} analyzes the role of model capacity. All methods are trained with SGD with $\eta = 0.01$ (decayed $\gamma = 0.995$/round), momentum $0.9$, weight decay $5 \times 10^{-4}$, gradient clipping at norm $1.0$, and $E = 5$ local epochs. For Fed-BAC, cluster parameters $\Theta_k$ receive additional L2 regularization ($\lambda = 0.001$) as described in Section~\ref{sec:additive}.

\subsubsection{Baselines and Configurations}
Three methods are compared:
\begin{enumerate}
    \item \textbf{HierFAVG} ($K\!=\!1$, $p\!=\!1.0$)~\cite{liu2020hierfavg}: Single global model with full participation. Baseline without personalization.
    \item \textbf{IFCA} ($K\!=\!5$, $p\!=\!1.0$)~\cite{ghosh2020ifca}: Isolated cluster models with loss-based reassignment ($\tau_{\text{re}} = 20$, threshold $0.95$), full participation. $K\!=\!5$ follows Fed-CAM~\cite{ma2023fedcam}, which demonstrated that IFCA is susceptible to imbalanced cluster assignment at higher $K$ values. Fed-BAC's $K_{\max}\!=\!M$ is an \textit{upper bound} on cluster granularity rather than a fixed partition, and the additive decomposition ensures all $M$ servers contribute to $\Theta_{\mathrm{global}}$ every round.
    \item \textbf{Fed-BAC} ($K_{\max}\!=\!M$, $p\!=\!0.8$): Additive clustering with LinUCB assignment and TS client selection. The LinUCB bandit determines the effective number of active clusters from the data. Cluster assignments are initialized via round-robin on CIFAR-10 and SVHN, and uniformly (all servers in one cluster) on Fashion-MNIST, to evaluate bandit discovery from both extremes.
\end{enumerate}

\subsubsection{Evaluation Metrics}
Following Fed-CAM~\cite{ma2023fedcam}, distributed accuracy is the primary metric: each server's cluster model is evaluated on its local test partition (partitioned using the same Dirichlet-sampled proportions as training data), and results are averaged across servers. This metric measures personalized performance under each server's local distribution rather than generalization to the global IID test set.

\begin{figure*}[t]
\centering
\begin{minipage}{0.32\textwidth}
\centering
\begin{tikzpicture}
\begin{axis}[
    width=\textwidth,
    height=4.2cm,
    xlabel={Round},
    ylabel={Dist.\ Accuracy (\%)},
    xmin=0, xmax=200,
    ymin=0, ymax=80,
    legend style={at={(0.98,0.02)}, anchor=south east, font=\tiny, draw=none, fill=white, fill opacity=0.8},
    grid=major,
    grid style={gray!20},
    tick label style={font=\tiny},
    label style={font=\scriptsize},
    title={\scriptsize (a) CIFAR-10, $\alpha\!=\!0.5$},
    title style={yshift=-3pt},
]
\addplot[green!60!black, thick, mark=none] coordinates {
  (1,9.4) (10,24.8) (20,36.4) (30,41.2) (40,44.0) (50,46.2) (60,48.0) (70,49.7)
  (80,51.2) (90,52.3) (100,53.2) (110,53.9) (120,54.5) (130,55.0) (140,55.5)
  (150,55.8) (160,56.2) (170,56.5) (180,56.8) (190,57.2) (200,57.5)
};
\addlegendentry{HierFAVG}
\addplot[red, thick, dashed, mark=none] coordinates {
  (1,8.9) (10,31.0) (20,43.3) (30,50.8) (40,53.9) (50,56.4) (60,57.8) (70,59.3)
  (80,60.3) (90,61.2) (100,61.9) (110,62.4) (120,63.0) (130,63.4) (140,63.7)
  (150,64.1) (160,64.4) (170,64.7) (180,64.9) (190,65.1) (200,65.3)
};
\addlegendentry{IFCA}
\addplot[blue, thick, mark=none] coordinates {
  (1,12.0) (10,42.3) (20,57.3) (30,63.4) (40,66.2) (50,67.9) (60,69.1) (70,69.9)
  (80,70.6) (90,71.2) (100,71.6) (110,72.0) (120,72.3) (130,72.6) (140,72.9)
  (150,73.0) (160,73.3) (170,73.4) (180,73.5) (190,73.7) (200,73.7)
};
\addlegendentry{Fed-BAC}
\end{axis}
\end{tikzpicture}
\end{minipage}%
\hfill
\begin{minipage}{0.32\textwidth}
\centering
\begin{tikzpicture}
\begin{axis}[
    width=\textwidth,
    height=4.2cm,
    xlabel={Round},
    xmin=0, xmax=150,
    ymin=0, ymax=95,
    legend style={at={(0.98,0.02)}, anchor=south east, font=\tiny, draw=none, fill=white, fill opacity=0.8},
    grid=major,
    grid style={gray!20},
    tick label style={font=\tiny},
    label style={font=\scriptsize},
    title={\scriptsize (b) SVHN, $\alpha\!=\!0.5$},
    title style={yshift=-3pt},
]
\addplot[green!60!black, thick, mark=none] coordinates {
  (1,8.0) (8,17.4) (15,36.4) (23,56.2) (30,66.5) (38,72.7) (45,75.7) (53,78.0)
  (60,79.4) (68,80.6) (75,81.3) (82,81.9) (90,82.5) (97,83.0) (105,83.4) (112,83.8)
  (120,84.1) (127,84.3) (135,84.5) (142,84.7) (150,84.9)
};
\addlegendentry{HierFAVG}
\addplot[red, thick, dashed, mark=none] coordinates {
  (1,8.0) (8,25.7) (15,44.2) (23,62.3) (30,72.8) (38,78.9) (45,81.7) (53,83.6)
  (60,84.6) (68,85.6) (75,86.1) (82,86.5) (90,86.9) (97,87.0) (105,87.3) (112,87.4)
  (120,87.6) (127,87.7) (135,87.8) (142,87.9) (150,88.0)
};
\addlegendentry{IFCA}
\addplot[blue, thick, mark=none] coordinates {
  (1,10.8) (8,33.6) (15,56.1) (23,71.9) (30,78.5) (38,82.5) (45,84.4) (53,85.8)
  (60,86.5) (68,87.1) (75,87.4) (82,87.8) (90,88.0) (97,88.1) (105,88.3) (112,88.4)
  (120,88.4) (127,88.5) (135,88.6) (142,88.6) (150,88.7)
};
\addlegendentry{Fed-BAC}
\end{axis}
\end{tikzpicture}
\end{minipage}%
\hfill
\begin{minipage}{0.32\textwidth}
\centering
\begin{tikzpicture}
\begin{axis}[
    width=\textwidth,
    height=4.2cm,
    xlabel={Round},
    xmin=0, xmax=100,
    ymin=0, ymax=95,
    legend style={at={(0.98,0.02)}, anchor=south east, font=\tiny, draw=none, fill=white, fill opacity=0.8},
    grid=major,
    grid style={gray!20},
    tick label style={font=\tiny},
    label style={font=\scriptsize},
    title={\scriptsize (c) Fashion-MNIST, $\alpha\!=\!0.5$},
    title style={yshift=-3pt},
]
\addplot[green!60!black, thick, mark=none] coordinates {
  (1,10.5) (5,28.3) (10,46.8) (15,57.8) (20,64.0) (25,67.7) (30,70.1) (35,71.8)
  (40,73.1) (45,74.2) (50,75.0) (55,75.8) (60,76.6) (65,77.1) (70,77.6) (75,78.1)
  (80,78.6) (85,79.0) (90,79.4) (95,79.7) (100,80.1)
};
\addlegendentry{HierFAVG}
\addplot[red, thick, dashed, mark=none] coordinates {
  (1,10.5) (5,32.0) (10,52.8) (15,64.6) (20,71.2) (25,75.8) (30,78.5) (35,80.0)
  (40,80.9) (45,81.5) (50,82.0) (55,82.4) (60,82.7) (65,82.9) (70,83.2) (75,83.4)
  (80,83.7) (85,83.9) (90,84.1) (95,84.4) (100,84.6)
};
\addlegendentry{IFCA}
\addplot[blue, thick, mark=none] coordinates {
  (1,10.6) (5,24.7) (10,44.7) (15,56.8) (20,64.6) (25,69.2) (30,73.0) (35,75.1)
  (40,77.6) (45,79.1) (50,79.8) (55,81.4) (60,82.6) (65,83.8) (70,84.8) (75,85.1)
  (80,85.5) (85,85.9) (90,86.2) (95,86.5) (100,86.7)
};
\addlegendentry{Fed-BAC}
\end{axis}
\end{tikzpicture}
\end{minipage}

\vspace{4pt}
\begin{minipage}{0.32\textwidth}
\centering
\begin{tikzpicture}
\begin{axis}[
    width=\textwidth,
    height=4.2cm,
    xlabel={Round},
    ylabel={Dist.\ Accuracy (\%)},
    xmin=0, xmax=200,
    ymin=0, ymax=95,
    legend style={at={(0.98,0.02)}, anchor=south east, font=\tiny, draw=none, fill=white, fill opacity=0.8},
    grid=major,
    grid style={gray!20},
    tick label style={font=\tiny},
    label style={font=\scriptsize},
    title={\scriptsize (d) CIFAR-10, $\alpha\!=\!0.1$},
    title style={yshift=-3pt},
]
\addplot[green!60!black, thick, mark=none] coordinates {
  (1,8.2) (10,16.9) (20,25.6) (30,30.9) (40,34.3) (50,36.8) (60,38.9) (70,40.8)
  (80,42.3) (90,43.5) (100,44.7) (110,45.5) (120,46.2) (130,46.7) (140,47.2)
  (150,47.7) (160,48.0) (170,48.5) (180,48.9) (190,49.2) (200,49.3)
};
\addlegendentry{HierFAVG}
\addplot[red, thick, dashed, mark=none] coordinates {
  (1,8.2) (10,39.1) (20,51.5) (30,66.5) (40,70.7) (50,73.1) (60,74.1) (70,74.8)
  (80,75.2) (90,75.4) (100,75.8) (110,75.9) (120,76.1) (130,76.3) (140,76.4)
  (150,76.5) (160,76.7) (170,76.8) (180,76.8) (190,76.9) (200,77.0)
};
\addlegendentry{IFCA}
\addplot[blue, thick, mark=none] coordinates {
  (1,10.1) (10,53.6) (20,71.9) (30,77.6) (40,80.2) (50,81.5) (60,82.1) (70,82.5)
  (80,82.8) (90,83.2) (100,83.5) (110,83.6) (120,83.7) (130,83.9) (140,84.1)
  (150,84.1) (160,84.1) (170,84.3) (180,84.4) (190,84.5) (200,84.9)
};
\addlegendentry{Fed-BAC}
\end{axis}
\end{tikzpicture}
\end{minipage}%
\hfill
\begin{minipage}{0.32\textwidth}
\centering
\begin{tikzpicture}
\begin{axis}[
    width=\textwidth,
    height=4.2cm,
    xlabel={Round},
    xmin=0, xmax=150,
    ymin=0, ymax=95,
    legend style={at={(0.98,0.02)}, anchor=south east, font=\tiny, draw=none, fill=white, fill opacity=0.8},
    grid=major,
    grid style={gray!20},
    tick label style={font=\tiny},
    label style={font=\scriptsize},
    title={\scriptsize (e) SVHN, $\alpha\!=\!0.1$},
    title style={yshift=-3pt},
]
\addplot[green!60!black, thick, mark=none] coordinates {
  (1,9.9) (8,15.7) (15,28.5) (23,44.2) (30,54.7) (38,63.3) (45,68.2) (53,71.7)
  (60,73.6) (68,75.0) (75,76.0) (82,76.7) (90,77.5) (97,78.0) (105,78.6) (112,79.0)
  (120,79.3) (127,79.6) (135,79.9) (142,80.1) (150,80.4)
};
\addlegendentry{HierFAVG}
\addplot[red, thick, dashed, mark=none] coordinates {
  (1,9.9) (8,31.1) (15,47.6) (23,64.0) (30,75.4) (38,81.2) (45,83.6) (53,85.0)
  (60,85.8) (68,86.5) (75,87.0) (82,87.4) (90,87.7) (97,87.9) (105,88.1) (112,88.2)
  (120,88.4) (127,88.5) (135,88.6) (142,88.7) (150,88.9)
};
\addlegendentry{IFCA}
\addplot[blue, thick, mark=none] coordinates {
  (1,9.3) (8,47.5) (15,68.5) (23,80.3) (30,85.6) (38,88.5) (45,89.7) (53,90.7)
  (60,91.3) (68,91.6) (75,91.9) (82,92.1) (90,92.4) (97,92.6) (105,92.5) (112,92.7)
  (120,92.7) (127,92.8) (135,92.8) (142,92.8) (150,92.9)
};
\addlegendentry{Fed-BAC}
\end{axis}
\end{tikzpicture}
\end{minipage}%
\hfill
\begin{minipage}{0.32\textwidth}
\centering
\begin{tikzpicture}
\begin{axis}[
    width=\textwidth,
    height=4.2cm,
    xlabel={Round},
    xmin=0, xmax=100,
    ymin=0, ymax=95,
    legend style={at={(0.98,0.02)}, anchor=south east, font=\tiny, draw=none, fill=white, fill opacity=0.8},
    grid=major,
    grid style={gray!20},
    tick label style={font=\tiny},
    label style={font=\scriptsize},
    title={\scriptsize (f) Fashion-MNIST, $\alpha\!=\!0.1$},
    title style={yshift=-3pt},
]
\addplot[green!60!black, thick, mark=none] coordinates {
  (1,8.2) (5,19.5) (10,37.1) (15,49.0) (20,56.4) (25,61.2) (30,64.4) (35,66.5)
  (40,67.9) (45,68.9) (50,69.6) (55,70.2) (60,70.7) (65,71.1) (70,71.4) (75,71.8)
  (80,72.1) (85,72.4) (90,72.7) (95,73.0) (100,73.3)
};
\addlegendentry{HierFAVG}
\addplot[red, thick, dashed, mark=none] coordinates {
  (1,7.5) (5,28.0) (10,66.8) (15,72.5) (20,75.5) (25,80.0) (30,85.6) (35,86.1)
  (40,86.5) (45,87.3) (50,87.9) (55,88.1) (60,88.3) (65,88.5) (70,88.8) (75,88.8)
  (80,88.8) (85,89.0) (90,89.4) (95,89.5) (100,89.7)
};
\addlegendentry{IFCA}
\addplot[blue, thick, mark=none] coordinates {
  (1,6.1) (5,39.3) (10,65.0) (15,78.6) (20,85.4) (25,89.0) (30,90.8) (35,91.8)
  (40,92.5) (45,92.9) (50,93.2) (55,93.4) (60,93.6) (65,93.8) (70,94.0) (75,94.1)
  (80,94.2) (85,94.3) (90,94.4) (95,94.5) (100,94.5)
};
\addlegendentry{Fed-BAC}
\end{axis}
\end{tikzpicture}
\end{minipage}
\caption{Distributed accuracy vs.\ communication rounds (exponentially smoothed). Top row: moderate heterogeneity ($\alpha\!=\!0.5$); bottom row: severe heterogeneity ($\alpha\!=\!0.1$). The gap between Fed-BAC and baselines widens at lower~$\alpha$.}
\label{fig:convergence}
\end{figure*}
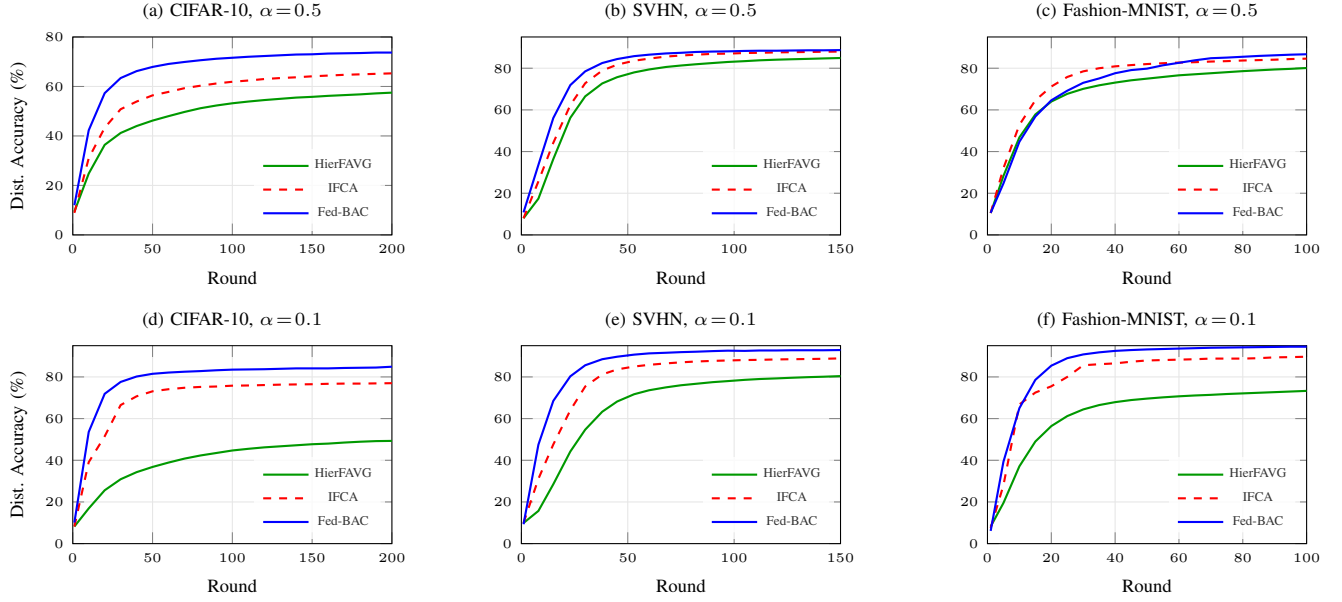

\subsection{Results}
\label{sec:results}

Table~\ref{tab:main_results} presents final distributed accuracy across all three datasets under both heterogeneity settings, and Table~\ref{tab:scaling} reports the corresponding evaluation at $5\times$ deployment scale on CIFAR-10.

\begin{table*}[t]
\centering
\caption{Distributed accuracy (\%) under moderate ($\alpha\!=\!0.5$) and severe ($\alpha\!=\!0.1$) heterogeneity. }
\label{tab:main_results}
\setlength{\tabcolsep}{3pt}
\begin{tabular}{lcc cc cc cc}
\toprule
& & & \multicolumn{2}{c}{\textbf{CIFAR-10} (200 rds)} & \multicolumn{2}{c}{\textbf{SVHN} (150 rds)} & \multicolumn{2}{c}{\textbf{Fashion-MNIST} (100 rds)} \\
\cmidrule(lr){4-5} \cmidrule(lr){6-7} \cmidrule(lr){8-9}
\textbf{Method} & $K$ & $p$ & $\alpha\!=\!0.5$ & $\alpha\!=\!0.1$ & $\alpha\!=\!0.5$ & $\alpha\!=\!0.1$ & $\alpha\!=\!0.5$ & $\alpha\!=\!0.1$ \\
\midrule
HierFAVG       & 1  & 1.0 & 57.53 $\pm$ 0.13 & 49.33 $\pm$ 0.16 & 84.93 $\pm$ 0.08 & 80.41 $\pm$ 0.19 & 80.13 $\pm$ 0.21 & 73.33 $\pm$ 0.19 \\
IFCA             & 5  & 1.0 & 65.31 $\pm$ 0.09 & 77.04 $\pm$ 0.10 & 88.01 $\pm$ 0.04 & 88.85 $\pm$ 0.13 & 84.65 $\pm$ 0.13 & 89.45 $\pm$ 1.34 \\
Fed-BAC           & $\leq\!M$ & 0.8 & 73.71 $\pm$ 0.11 & 84.87 $\pm$ 0.35 & 88.67 $\pm$ 0.06 & 92.90 $\pm$ 0.12 & 86.66 $\pm$ 0.84 & 94.54 $\pm$ 0.13 \\
\midrule
\multicolumn{3}{l}{$\Delta_{\text{H}}$ (Fed-BAC $-$ HierFAVG)} & +16.2pp & +35.5pp & +3.7pp & +12.5pp & +6.5pp & +21.2pp \\
\multicolumn{3}{l}{$\Delta_{\text{I}}$ (Fed-BAC $-$ IFCA)} & +8.4pp & +7.8pp & +0.7pp & +4.1pp & +2.0pp & +5.1pp \\
\bottomrule
\end{tabular}
\end{table*}

The analysis highlights four key results.

\textit{Fed-BAC's advantage increases with heterogeneity.} Under severe heterogeneity ($\alpha\!=\!0.1$), HierFAVG degrades substantially (57.53\% $\to$ 49.33\% on CIFAR-10, 80.13\% $\to$ 73.33\% on Fashion-MNIST), while both IFCA and Fed-BAC improve. Fed-BAC's gap over HierFAVG widens from +16.2pp to +35.5pp on CIFAR-10, from +3.7pp to +12.5pp on SVHN, and from +6.5pp to +21.2pp on Fashion-MNIST. Fed-BAC also maintains its advantage over IFCA at $\alpha\!=\!0.1$ (+7.8pp on CIFAR-10, +4.1pp on SVHN, +5.1pp on Fashion-MNIST).

\textit{Participation-efficient clustering.} Despite operating at $p\!=\!0.8$, Fed-BAC matches or exceeds IFCA ($K\!=\!5$, $p\!=\!1.0$) while activating 20\% fewer clients per round, yielding $3.0\times$ to $5.9\times$ lower total communication to representative accuracy targets on CIFAR-10 (Section~\ref{sec:system_model}).

\textit{Gains hold at larger deployment scale.} Table~\ref{tab:scaling} reports CIFAR-10 results at $5\times$ client population. Fed-BAC retains its lead on distributed accuracy, exceeding HierFAVG by +12.6pp and IFCA by +3.3pp. The distributed accuracy of all three methods is lower than at the primary scale because per-client training volume is reduced by a factor of five under the same Dirichlet partition, which is a property of the data partition rather than of any individual method. The margin over IFCA compresses from +8.4pp at primary scale to +3.3pp at $5\times$ scale; the lead nonetheless remains positive, confirming that Fed-BAC's additive cluster personalization continues to provide a measurable benefit over isolated clustering at larger deployments.

\textit{Convergence speed scales gracefully.} The convergence advantage is preserved at larger scale. At round~20 of the $5\times$ CIFAR-10 run, Fed-BAC reaches 38.8\% distributed accuracy versus 25.2\% for HierFAVG and 29.6\% for IFCA, confirming that bandit-guided personalization retains its early-round efficiency as the client population grows.

\begin{table}[t]
\centering
\caption{Scaling study on CIFAR-10 ($M\!=\!20$, $N\!=\!500$, $\alpha\!=\!0.5$).}
\label{tab:scaling}
\setlength{\tabcolsep}{4pt}
\begin{tabular}{l c}
\toprule
\textbf{Method} & \textbf{Dist.\ Acc.\ (\%)} \\
\midrule
HierFAVG          & 39.97 $\pm$ 0.07 \\
IFCA ($K\!=\!5$)  & 49.33 $\pm$ 2.16 \\
Fed-BAC ($K_\leq\!M$ ) & 52.60 $\pm$ 1.84 \\
\bottomrule
\end{tabular}
\end{table}

\subsection{Convergence Speed}
\label{sec:convergence}

Fig.~\ref{fig:convergence} presents convergence trajectories under both heterogeneity settings. Convergence speed is quantified as the ratio $T_{\text{baseline}} / T_{\text{Fed-BAC}}$, where $T_{\text{method}}$ denotes the number of communication rounds required by a given method to first reach a specified distributed accuracy threshold; a ratio greater than one indicates that Fed-BAC reaches the target faster. Under moderate heterogeneity ($\alpha\!=\!0.5$, top row), Fed-BAC achieves 50\% distributed accuracy on CIFAR-10 by round 15 versus round 72 for HierFAVG (4.8$\times$), and reaches each SVHN target 1.5--1.9$\times$ faster than HierFAVG. HierFAVG and IFCA plateau below 58\% and 66\% on CIFAR-10, whereas Fed-BAC is the only method to exceed 70\%. On Fashion-MNIST, IFCA converges faster at lower targets due to its fixed $K\!=\!5$ clusters; Fed-BAC overtakes IFCA after approximately 60 rounds and is the only method to reach 85\%.

Under severe heterogeneity ($\alpha\!=\!0.1$, bottom row), the gap widens further: on CIFAR-10, Fed-BAC surpasses 80\% by round~40 while HierFAVG plateaus below 50\%, and on SVHN, Fed-BAC exceeds 90\% by round~53 versus round~142 for HierFAVG to reach only 80\%.

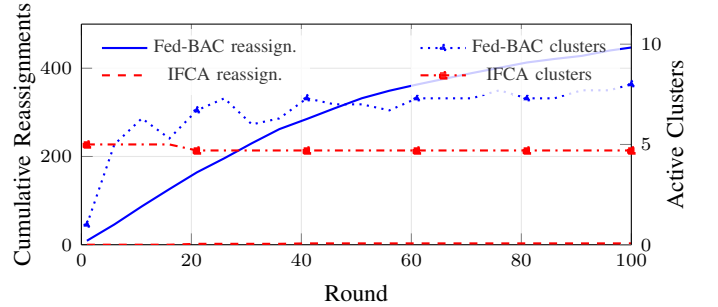
\begin{figure}[t]
\centering
\begin{tikzpicture}
\begin{axis}[
    width=\columnwidth,
    height=4.5cm,
    xlabel={Round},
    ylabel={Cumulative Reassignments},
    ylabel style={yshift=-5pt},
    xmin=0, xmax=100,
    ymin=0, ymax=500,
    legend style={at={(0.02,0.98)}, anchor=north west, font=\scriptsize, draw=none, fill=white, fill opacity=0.8},
    grid=major,
    grid style={gray!20},
    tick label style={font=\scriptsize},
    label style={font=\small},
    axis y line*=left,
]
\addplot[blue, thick, mark=none] coordinates {
  (1,9) (6,46) (11,87) (16,126) (21,164) (26,196) (31,230) (36,262)
  (41,285) (46,309) (51,332) (56,349) (61,363) (66,376) (71,389)
  (76,401) (81,413) (86,421) (91,428) (96,440) (100,447)
};
\addlegendentry{Fed-BAC reassign.}

\addplot[red, thick, dashed, mark=none] coordinates {
  (1,0) (6,0) (11,0) (16,0) (21,2) (26,2) (31,2) (36,2) (41,3)
  (46,3) (51,3) (56,3) (61,3) (66,3) (71,3) (76,3) (81,3) (86,3)
  (91,3) (96,3) (100,3)
};
\addlegendentry{IFCA reassign.}
\end{axis}

\begin{axis}[
    width=\columnwidth,
    height=4.5cm,
    xmin=0, xmax=100,
    ymin=0, ymax=11,
    ylabel={Active Clusters},
    ylabel style={yshift=5pt},
    axis y line*=right,
    axis x line=none,
    legend style={at={(0.98,0.98)}, anchor=north east, font=\scriptsize, draw=none, fill=white, fill opacity=0.8},
    tick label style={font=\scriptsize},
    label style={font=\small},
]
\addplot[blue, thick, dotted, mark=triangle*, mark size=1.5pt, mark repeat=4] coordinates {
  (1,1.0) (6,5.0) (11,6.3) (16,5.3) (21,6.7) (26,7.3) (31,6.0)
  (36,6.3) (41,7.3) (46,7.0) (51,7.0) (56,6.7) (61,7.3) (66,7.3)
  (71,7.3) (76,7.7) (81,7.3) (86,7.3) (91,7.7) (96,7.7) (100,8.0)
};
\addlegendentry{Fed-BAC clusters}

\addplot[red, thick, dashdotted, mark=square*, mark size=1.2pt, mark repeat=4] coordinates {
  (1,5.0) (6,5.0) (11,5.0) (16,5.0) (21,4.7) (26,4.7) (31,4.7)
  (36,4.7) (41,4.7) (46,4.7) (51,4.7) (56,4.7) (61,4.7) (66,4.7)
  (71,4.7) (76,4.7) (81,4.7) (86,4.7) (91,4.7) (96,4.7) (100,4.7)
};
\addlegendentry{IFCA clusters}
\end{axis}
\end{tikzpicture}
\caption{Cluster dynamics on Fashion-MNIST ($\alpha\!=\!0.5$, 100 rounds).}
\label{fig:cluster_dynamics_fmnist}
\end{figure}

\begin{figure}[t]
\centering
\begin{tikzpicture}
\begin{axis}[
    width=\columnwidth,
    height=4.5cm,
    xlabel={Round},
    ylabel={Cumulative Reassignments},
    ylabel style={yshift=-5pt},
    xmin=0, xmax=200,
    ymin=0, ymax=700,
    legend style={at={(0.02,0.98)}, anchor=north west, font=\scriptsize, draw=none, fill=white, fill opacity=0.8},
    grid=major,
    grid style={gray!20},
    tick label style={font=\scriptsize},
    label style={font=\small},
    axis y line*=left,
]
\addplot[blue, thick, mark=none] coordinates {
  (1,0) (10,74) (20,162) (30,241) (40,315) (50,379) (60,432) (70,470)
  (80,498) (90,517) (100,535) (110,557) (120,570) (130,586) (140,595)
  (150,604) (160,613) (170,622) (180,636) (190,639) (200,647)
};
\addlegendentry{Fed-BAC reassign.}

\addplot[red, thick, dashed, mark=none] coordinates {
  (1,0) (10,0) (20,0) (30,1) (40,1) (50,2) (60,2) (70,2) (80,2)
  (90,2) (100,2) (110,2) (120,2) (130,2) (140,2) (150,2) (160,2)
  (170,2) (180,2) (190,2) (200,2)
};
\addlegendentry{IFCA reassign.}
\end{axis}

\begin{axis}[
    width=\columnwidth,
    height=4.5cm,
    xmin=0, xmax=200,
    ymin=0, ymax=11,
    ylabel={Active Clusters},
    ylabel style={yshift=5pt},
    axis y line*=right,
    axis x line=none,
    legend style={at={(0.98,0.98)}, anchor=north east, font=\scriptsize, draw=none, fill=white, fill opacity=0.8},
    tick label style={font=\scriptsize},
    label style={font=\small},
]
\addplot[blue, thick, dotted, mark=triangle*, mark size=1.5pt, mark repeat=4] coordinates {
  (1,10.0) (10,4.3) (20,7.0) (30,7.0) (40,6.7) (50,6.7) (60,7.0)
  (70,7.7) (80,7.3) (90,8.3) (100,7.7) (110,7.7) (120,8.3) (130,8.7)
  (140,7.7) (150,8.0) (160,8.7) (170,8.3) (180,8.7) (190,8.0) (200,8.0)
};
\addlegendentry{Fed-BAC clusters}

\addplot[red, thick, dashdotted, mark=square*, mark size=1.2pt, mark repeat=4] coordinates {
  (1,5.0) (10,5.0) (20,5.0) (30,5.0) (40,5.0) (50,5.0) (60,5.0)
  (70,5.0) (80,5.0) (90,5.0) (100,5.0) (110,5.0) (120,5.0) (130,5.0)
  (140,5.0) (150,5.0) (160,5.0) (170,5.0) (180,5.0) (190,5.0) (200,5.0)
};
\addlegendentry{IFCA clusters}
\end{axis}
\end{tikzpicture}
\caption{Cluster dynamics on CIFAR-10 ($\alpha\!=\!0.5$, 200 rounds).}
\label{fig:cluster_dynamics_cifar10}
\end{figure}
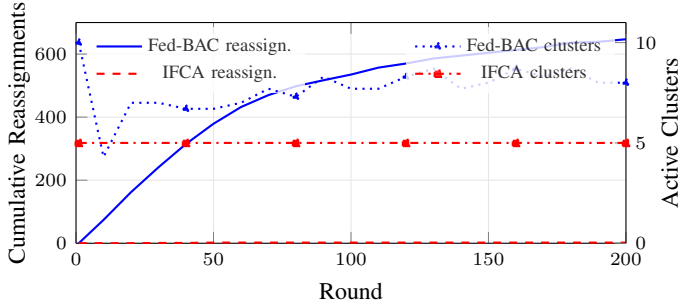

\begin{figure}[t]
\centering
\begin{tikzpicture}
\begin{axis}[
    width=\columnwidth,
    height=4.5cm,
    xlabel={Round},
    ylabel={Cumulative Reassignments},
    ylabel style={yshift=-5pt},
    xmin=0, xmax=150,
    ymin=0, ymax=700,
    legend style={at={(0.02,0.98)}, anchor=north west, font=\scriptsize, draw=none, fill=white, fill opacity=0.8},
    grid=major,
    grid style={gray!20},
    tick label style={font=\scriptsize},
    label style={font=\small},
    axis y line*=left,
]
\addplot[blue, thick, mark=none] coordinates {
  (1,0) (10,78) (20,166) (30,243) (40,298) (50,347) (60,397) (70,434)
  (80,469) (90,507) (100,536) (110,561) (120,584) (130,615) (140,627)
  (150,642)
};
\addlegendentry{Fed-BAC reassign.}

\addplot[red, thick, dashed, mark=none] coordinates {
  (1,0) (10,0) (20,0) (30,3) (40,3) (50,3) (60,3) (70,3) (80,3)
  (90,3) (100,3) (110,3) (120,3) (130,3) (140,3) (150,3)
};
\addlegendentry{IFCA reassign.}
\end{axis}

\begin{axis}[
    width=\columnwidth,
    height=4.5cm,
    xmin=0, xmax=150,
    ymin=0, ymax=11,
    ylabel={Active Clusters},
    ylabel style={yshift=5pt},
    axis y line*=right,
    axis x line=none,
    legend style={at={(0.98,0.98)}, anchor=north east, font=\scriptsize, draw=none, fill=white, fill opacity=0.8},
    tick label style={font=\scriptsize},
    label style={font=\small},
]
\addplot[blue, thick, dotted, mark=triangle*, mark size=1.5pt, mark repeat=3] coordinates {
  (1,10.0) (10,5.3) (20,6.7) (30,6.3) (40,7.3) (50,7.7) (60,7.0)
  (70,7.3) (80,7.7) (90,7.3) (100,7.7) (110,8.0) (120,8.3) (130,8.0)
  (140,8.3) (150,8.3)
};
\addlegendentry{Fed-BAC clusters}

\addplot[red, thick, dashdotted, mark=square*, mark size=1.2pt, mark repeat=3] coordinates {
  (1,5.0) (10,5.0) (20,5.0) (30,4.0) (40,4.0) (50,4.0) (60,4.0)
  (70,4.0) (80,4.0) (90,4.0) (100,4.0) (110,4.0) (120,4.0) (130,4.0)
  (140,4.0) (150,4.0)
};
\addlegendentry{IFCA clusters}
\end{axis}
\end{tikzpicture}
\caption{Cluster dynamics on SVHN ($\alpha\!=\!0.5$, 150 rounds).}
\label{fig:cluster_dynamics_svhn}
\end{figure}
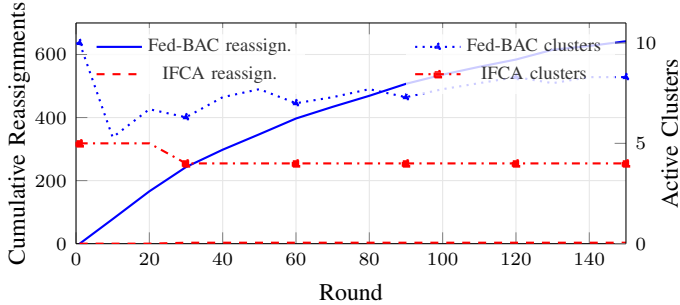

\begin{table*}[t]
\centering
\caption{Per-server distributed accuracy distribution (last 10 rounds). $\sigma$ denotes the cross-server standard deviation of accuracy.}
\label{tab:fairness}
\setlength{\tabcolsep}{3.5pt}
\begin{tabular}{cl cccc cccc}
\toprule
& & \multicolumn{4}{c}{$\alpha\!=\!0.5$} & \multicolumn{4}{c}{$\alpha\!=\!0.1$} \\
\cmidrule(lr){3-6} \cmidrule(lr){7-10}
\textbf{Dataset} & \textbf{Method} & \textbf{Mean} & \textbf{Min} & \textbf{Max} & $\boldsymbol{\sigma}$ & \textbf{Mean} & \textbf{Min} & \textbf{Max} & $\boldsymbol{\sigma}$ \\
\midrule
\multirow{3}{*}{\rotatebox[origin=c]{90}{{\scriptsize CIFAR-10}}}
& HierFAVG                       & 57.53 & 53.04 & 59.71 & 2.10 & 49.33 & 37.72 & 60.03 & 6.38 \\
& IFCA $K\!=\!5$                   & 65.31 & 53.28 & 72.15 & 5.27          & 77.04 & 72.01 & 88.34 & 5.57 \\
& Fed-BAC $K_{\max}\!=\!M$        & 73.71 & 69.54 & 76.94 & 2.78 & 84.87 & 80.43 & 93.62 & 3.96 \\
\midrule
\multirow{3}{*}{\rotatebox[origin=c]{90}{\scriptsize SVHN}}
& HierFAVG                       & 84.93 & 76.74 & 88.03 & 3.19          & 80.41 & 75.51 & 84.84 & 2.78 \\
& IFCA $K\!=\!5$                   & 88.01 & 85.32 & 91.44 & 1.97          & 88.85 & 79.98 & 95.62 & 4.79 \\
& Fed-BAC $K_{\max}\!=\!M$        & 88.67 & 86.40 & 90.92 & 1.41 & 92.90 & 87.89 & 96.08 & 2.17 \\
\midrule
\multirow{3}{*}{\rotatebox[origin=c]{90}{\parbox{1cm}{\centering\scriptsize Fashion-\\[-2pt]MNIST}}}
& HierFAVG                       & 80.13 & 71.90 & 84.89 & 4.64          & 73.33 & 56.83 & 86.72 & 9.22 \\
& IFCA $K\!=\!5$                   & 84.65 & 75.56 & 90.45 & 4.66          & 89.45 & 82.82 & 94.68 & 3.98 \\
& Fed-BAC $K_{\max}\!=\!M$        & 86.66 & 83.79 & 89.71 & 2.05 & 94.54 & 90.16 & 99.06 & 2.45 \\
\bottomrule
\end{tabular}
\end{table*}

\subsection{Discussion}
\label{sec:discussion}

\textit{Why clustering methods improve at higher heterogeneity.} IFCA and Fed-BAC achieve \textit{higher} distributed accuracy at $\alpha\!=\!0.1$ than at $\alpha\!=\!0.5$, while HierFAVG degrades. Under severe heterogeneity, server distributions become more distinct, producing cleaner cluster structure that clustering methods exploit, whereas HierFAVG's single global model generalizes poorly to any individual server.

\textit{Knowledge sharing through additive decomposition.} IFCA's isolated clusters cannot share knowledge even when distributions partially overlap. Fed-BAC's additive formulation addresses this by routing all server contributions through $\Theta_{\mathrm{global}}$, and the consistent accuracy gap between Fed-BAC and IFCA across both heterogeneity levels (Table~\ref{tab:main_results}) demonstrates that the shared global component captures transferable structure that isolated clustering discards.

\textit{Cluster dynamics.} Fed-BAC's LinUCB bandit reassigns servers actively in early rounds and stabilizes as the bandit converges (Figs.~\ref{fig:cluster_dynamics_fmnist}--\ref{fig:cluster_dynamics_svhn}). On Fashion-MNIST (initialized as one cluster), Fed-BAC discovers 7--8 active clusters versus IFCA's static 3; on CIFAR-10 and SVHN (initialized with $M$ distinct clusters), Fed-BAC stabilizes at ${\approx}8$ clusters, indicating robust discovery from both initialization extremes.

\textit{Cross-server fairness.} Table~\ref{tab:fairness} reports per-server accuracy distributions. At $\alpha\!=\!0.5$, Fed-BAC achieves the lowest cross-server variance on SVHN and Fashion-MNIST ($\sigma\!=\!1.41$ and $2.05$) and halves IFCA's variance on CIFAR-10, while its worst server (69.54\%) exceeds every other method's mean. Under severe heterogeneity ($\alpha\!=\!0.1$), HierFAVG's $\sigma$ increases sharply (2.10~$\to$~6.38 on CIFAR-10, 4.64~$\to$~9.22 on Fashion-MNIST), whereas Fed-BAC limits degradation ($\sigma\!=\!3.96$ and $2.45$), with its worst CIFAR-10 server (80.43\%) still exceeding IFCA's mean (77.04\%).

\textit{Role of model capacity.} Fed-BAC uses ${\approx}124$K total parameters versus ${\approx}62$K for baselines. To isolate the effect of capacity, HierFAVG is evaluated with a wider LeNet-5 matching Fed-BAC's total parameter count (${\approx}124$K, $p\!=\!1.0$). On CIFAR-10 at $\alpha\!=\!0.5$, HierFAVG~2$\times$ achieves 65.02\% distributed accuracy, a +7.5pp improvement over the standard 62K model (57.53\%) but still 8.7pp below Fed-BAC (73.71\%) at the same capacity and lower participation. Capacity therefore accounts for less than half the total gap; the remainder is attributable to the additive decomposition and bandit-guided decisions. Notably, HierFAVG~2$\times$ (65.02\%) merely matches IFCA at 62K (65.31\%), confirming that a wider single global model cannot substitute for cluster personalization.

\begin{table}[t]
\centering
\caption{TS vs.\ random client selection ($p\!=\!0.8$, $K_{\max}\!=\!M$, $\alpha\!=\!0.5$).}
\label{tab:client_selection}
\setlength{\tabcolsep}{4pt}
\begin{tabular}{cl cc}
\toprule
\textbf{Dataset} & \textbf{Selection} & \textbf{Dist.\ Acc.\ (\%)} & $\boldsymbol{\sigma}_{\text{server}}$ \\
\midrule
\multirow{2}{*}{CIFAR-10}
& Random  & 70.95 $\pm$ 2.44 & 3.23 \\
& TS      & 73.71 $\pm$ 1.03 & 2.78 \\
\midrule
\multirow{2}{*}{\parbox{1.2cm}{\centering Fashion-\\[-2pt]MNIST}}
& Random  & 85.16 $\pm$ 1.40 & 3.55 \\
& TS      & 86.66 $\pm$ 2.16 & 2.05 \\
\bottomrule
\end{tabular}
\end{table}

\subsection{Impact of Client Selection Strategy}
\label{sec:client_selection}

To isolate the contribution of TS, Fed-BAC with TS-based selection ($p\!=\!0.8$) is compared against uniform random selection at the same participation rate (Table~\ref{tab:client_selection}). Since both variants use the same additive architecture (${\approx}\,124$K parameters), this also controls for the $2\times$ capacity difference relative to baselines. At this scale, TS selects 8 of 10 clients per round, so posteriors diverge slowly. The resulting gains are modest ($+2.8$pp on CIFAR-10, $+1.5$pp on Fashion-MNIST), though TS consistently reduces cross-server variance ($\sigma$: $3.55 \to 2.05$ on Fashion-MNIST) at zero additional communication cost. Larger gains are expected at lower participation rates. 

\section{Conclusion}
\label{sec:conclusion}

This work presented Fed-BAC, an HFL framework that jointly optimizes cluster assignment and client selection in three-tier MEC architectures through a two-level bandit framework integrated with additive cluster personalization. Across three benchmarks at the primary configuration (100 clients, 10 edge servers) and a $5\times$ scaling study on CIFAR-10 (500 clients, 20 edge servers), Fed-BAC consistently outperforms both HierFAVG and IFCA in distributed accuracy while operating at only 80\% client participation, with the advantage widening under more severe data heterogeneity. The LinUCB bandit reliably discovers cluster structure from both initialization extremes, and TS improves both accuracy and cross-server fairness without additional communication overhead.

The current evaluation uses a single participation rate ($p\!=\!0.8$); a systematic sweep over $p \in \{0.6, 0.7, 0.8, 0.9\}$ would characterize the accuracy-participation trade-off. Natural extensions include real-world testbed deployments on edge hardware, evaluation under more challenging scenarios such as client dropout, asynchronous participation, and adversarial conditions, and formal convergence guarantees for the coupled non-stationary bandit system.

\bibliographystyle{ieeetr}
\bibliography{ref}

\end{document}